\documentclass[letterpaper, 10 pt, journal, twoside]{ieeetran}   

\usepackage{array}
\usepackage{amsmath}
\usepackage{amssymb}
\usepackage{booktabs}
\usepackage{cite}
\usepackage{graphics}
\usepackage{graphicx}
\usepackage{hyperref}
\usepackage{multirow}
\usepackage{tabularx}
\usepackage{times}
\usepackage{xcolor}
\usepackage[ruled,vlined,linesnumbered]{algorithm2e}
\usepackage{threeparttable}

\begin{document}
\title{Communication-Efficient Module-Wise Federated Learning for \\ Grasp Pose Detection in Cluttered Environments}

\author{Woonsang Kang$^{1}$, Joohyung Lee$^{2}$, Seungjun Kim$^{3}$, Jungchan Cho$^{2*}$, Yoonseon Oh$^{1*}$
\thanks{Manuscript received: July 7, 2025; October 5, 2025; Accepted November 4, 2025.}
\thanks{This paper was recommended for publication by Editor Borràs Sol Júlia upon evaluation of the Associate Editor and Reviewers' comments.

This work was partly (50\%) supported by the National Research Foundation of Korea (NRF) grant funded by the Korea government (MSIT) (No. RS-2024-00409492, Beyond-G Global Innovation Center).

This work was partly (50\%) supported by the National Research Foundation of Korea(NRF) grant funded by the Korea government(MSIT) (RS-2025-22442984).
(Co-corresponding authors: Yoonseon Oh and Jungchan Cho.)
}
\thanks{$^{1}$Woonsang Kang and Yoonseon Oh are with the Department of Electronic Engineering, 
Hanyang University, Seoul, South Korea  {\tt\footnotesize monni1729 @hanyang.ac.kr; yoh21@hanyang.ac.kr}}
\thanks{$^{2}$Joohyung Lee and Jungchan Cho are with the Department of Computing, 
Gachon University, Seongnam, South Korea {\tt\footnotesize j17.lee@gachon.ac.kr; thinkai@gachon.ac.kr}}
\thanks{$^{3}$Seungjun Kim is with the Department of Artificial Intelligence, 
Hanyang University, Seoul, South Korea {\tt\footnotesize rlatmdwnseo@gachon.ac.kr}}
\thanks{Digital Object Identifier (DOI): see top of this page.}}
\markboth{IEEE ROBOTICS AND AUTOMATION LETTERS. PREPRINT VERSION. ACCEPTED NOVEMBER, 2025}
{Kang \MakeLowercase{\textit{et al.}}: Communication-Efficient Federated Learning for Grasp Pose Detection} 
\maketitle
\begin{abstract}
Grasp pose detection (GPD) is a fundamental capability for robotic autonomy, but its reliance on large, diverse datasets creates significant data privacy and centralization challenges. Federated Learning (FL) offers a privacy-preserving solution, but its application to GPD is hindered by the substantial communication overhead of large models, a key issue for resource-constrained robots. To address this, we propose a novel module-wise FL framework that begins by analyzing the learning dynamics of the GPD model's functional components. This analysis identifies slower-converging modules, to which our framework then allocates additional communication effort. This is realized through a two-phase process: a standard full-model training phase is followed by a communication-efficient phase where only an adaptively identified subset of slower-converging modules is trained and their partial updates are aggregated. Extensive experiments on the GraspNet-1B dataset demonstrate that our method outperforms standard FedAvg and other baselines, achieving higher accuracy for a given communication budget. Furthermore, real-world experiments on a physical robot validate our approach, showing a superior grasp success rate compared to baseline methods in cluttered scenes. Our work presents a communication-efficient framework for training robust, generalized GPD models in a decentralized manner, effectively improving the trade-off between communication cost and model performance.
\end{abstract}


\section{Introduction}
\IEEEPARstart{G}{rasp} pose detection (GPD) is a fundamental capability for autonomous robotic systems, with applications ranging from industrial automation to assistive robotics. In recent years, learning-based approaches have become the dominant paradigm for this task~\cite{bohg2013data, fang2020graspnet}. This success, however, is critically dependent on large-scale, diverse datasets capable of fostering generalization across various objects, scenes, and environmental conditions~\cite{mahler2016dex, levine2018learning, fang2020graspnet, vuong2024grasp}.

However, acquiring such data presents substantial challenges. The collection process is not only laborious but, more critically, raises significant privacy concerns when data originates from private settings like homes or proprietary industrial environments~\cite{pinto2016supersizing, fang2020graspnet, mcmahan2017communication}. The traditional requirement of centralizing sensitive sensor information thus becomes a major obstacle, restricting the ability to pool datasets from multiple sources. This limitation ultimately hinders the training of more robust, generalized models and creates a \textit{data island} problem that impedes the development of universally applicable GPD.

\begin{figure}[t]
\centering
\includegraphics[width=0.9\columnwidth]{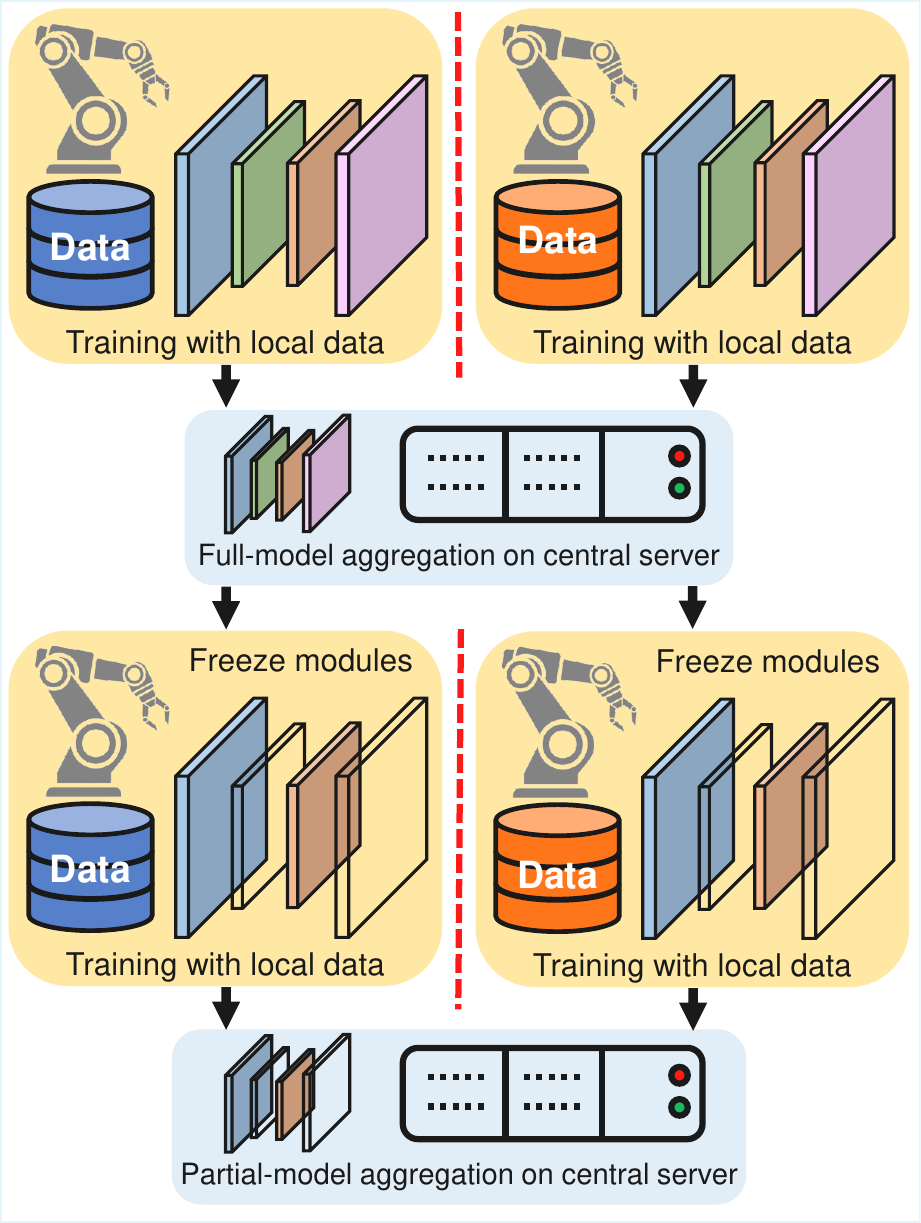}
\caption{An overview of our proposed two-phase module-wise FL algorithm. Phase 1 consists of standard full-model training and aggregation. In contrast, Phase 2 enhances communication efficiency with a partial-update strategy: based on our module-wise analysis, the faster-converging modules (visualized as transparent plates) are frozen. Only the remaining, slower-converging modules are trained, and their partial updates are aggregated by the server. The red dotted lines represent the privacy boundary, ensuring no raw data is shared. The aggregation step involves sharing only model weights.}
\label{fig:overview}
\end{figure}

Federated Learning (FL) has emerged as a compelling paradigm for addressing privacy and data governance challenges, enabling multiple clients to collaboratively train a model without exchanging their raw data~\cite{mcmahan2017communication}. This decentralized approach is highly pertinent to robotics, allowing for the use of diverse, distributed datasets for applications like autonomous navigation~\cite{gummadi2024fed} and manipulator control \cite{wang2024grasp} while preserving data privacy. However, a significant bottleneck in FL is the \textit{communication} overhead, particularly for the large-parameter models used in GPD~\cite{mcmahan2017communication}. The frequent transmission of model updates consumes substantial bandwidth and energy, a critical issue for resource-constrained robots. Although prior works like~\cite{kang2023fogl} have aimed to address data heterogeneity by clustering robots (or clients), their approach can intensify this resource burden.

To address the resource burden of FL for GPD, we observe that contemporary models in this domain feature \textit{multi-modular} architectures~\cite{fang2020graspnet, wang2021graspness, wang2023granet, wu2024economic}. To analyze these architectures in an FL setting, we measured the cosine similarity of model updates on a module-by-module basis~\cite{sattler2020clustered, ma2022convergence}. Our analysis reveals that these modules exhibit heterogeneous learning dynamics, allowing us to distinguish between faster-converging modules and the slower-converging ones that act as training bottlenecks. This finding motivates our proposal of a novel, two-phase module-wise FL algorithm, as illustrated in Figure~\ref{fig:overview}. The first phase consists of standard full-model training and full aggregation. The second phase, in contrast, employs a parameter-efficient approach by strategically concentrating resources: only the subset of modules identified as slower-converging undergoes partial-model training, and subsequently, only their updated parameters are aggregated. This approach maximizes the performance return on a given communication budget, enhancing the feasibility of FL for complex robotics tasks like GPD. Our primary contributions are as follows:
\begin{itemize}
    \item We identify communication overhead as a critical bottleneck for FL in GPD. We introduce a module-wise similarity analysis that reveals heterogeneous learning dynamics and, crucially, enables the identification of specific slower-converging modules that act as training bottlenecks.
    \item Motivated by our analysis, we propose a novel module-wise FL algorithm featuring a two-phase process. This algorithm allocates additional training and communication resources exclusively to the slower-converging modules adaptively, enhancing communication efficiency and maximizing performance for a given budget.
    \item Through extensive experiments on the GraspNet-1B dataset \cite{fang2020graspnet} and in the real world, we demonstrate the superiority of our approach over existing baselines and validate its practical effectiveness.
\end{itemize}


\section{Related work}
\subsection{Data-Driven Grasp Pose Detection}
GPD aims to identify a stable 6-DoF (Degree of freedom) gripper pose from sensory inputs like RGB-D images or 3D point clouds. Historically, this problem was often addressed through model-based approaches~\cite{bohg2013data}, which relied on having precise 3D models of the objects. However, these methods faced significant limitations, such as difficulty handling novel objects and sensitivity to sensor noise. The advent of deep learning has since spurred a paradigm shift towards data-driven grasping methodologies that learn directly from data. The field's progress is driven by advanced datasets \cite{mahler2016dex, fang2020graspnet, vuong2024grasp} and novel model architectures \cite{zhang2019roi, wang2023granet}. Early data-driven works focused on representing grasps as 2D rectangles in the image plane \cite{lenz2015deep, levine2018learning}. However, these 2D approaches could not represent the full spatial orientation required for manipulation in \textit{cluttered scenes}. This limitation spurred a paradigm shift towards methods that detect full 6-DoF poses directly from 3D point clouds. These modern approaches have progressed from sampling-based techniques that evaluate numerous candidates \cite{fang2020graspnet} to more efficient, single-stage frameworks that directly predict grasp poses \cite{zhang2019roi, liu2022transgrasp, ma2023towards, wang2023granet}. This advancement is strongly supported by the development of large-scale datasets, which are crucial for training and evaluation. These range from early synthetic datasets with millions of grasp examples \cite{mahler2016dex} to recent real-world benchmarks focusing on cluttered environments and object diversity \cite{fang2020graspnet, vuong2024grasp}. However, the very success built upon these diverse, real-world datasets underscores a critical challenge: the logistical and privacy-related difficulties in centralizing such large-scale, sensitive data.

\subsection{Federated Learning for Robotic Grasping}
FL has emerged as a machine learning paradigm that enables multiple decentralized clients to collaboratively train a model without sharing their raw local datasets~\cite{mcmahan2017communication}. In this framework, exemplified by the canonical FedAvg algorithm, clients locally train a shared model, and a central server aggregates only their resulting updates to produce an improved global model. The inherent privacy preservation of this framework makes it highly suitable for robotics, where applications often involve sensitive data from private residences or proprietary industrial processes. Reflecting its growing importance, FL is now being applied to a range of robotic tasks to scaling robot learning. It includes multi-robot trajectory prediction~\cite{majcherczyk2021flow}, visual robotic navigation~\cite{gummadi2024fed}, and robotic manipulation \cite{wang2024grasp}, \cite{miao2025fedvla}.

However, applying FL to GPD is still in its early stages. While a prior work by \cite{kang2023fogl} did explore this direction by applying a clustering-based FL approach to 2D grasping, it suffers from two critical limitations. First, from a task perspective, it is confined to simplified 2D rectangular grasps, which are insufficient for dexterous manipulation in cluttered 3D scenes. Second, from a methodological standpoint, its clustering process not only incurs significant communication overhead but also risks harming generalization by permanently excluding clients with unique data distributions. Therefore, applying FL to the more practical and challenging GPD problem, while simultaneously addressing the communication-efficiency of the FL algorithm itself, remains a significant open challenge. Our work aims to fill this critical gap by proposing a novel, communication-efficient FL strategy specifically designed for multi-modular GPD models.

\subsection{Communication-Efficient Federated Learning}
A primary bottleneck in FL is the substantial communication overhead required for the iterative transmission of large-scale models between clients and a central server. This challenge is exacerbated as model complexity and the number of participating clients increase, creating a critical trade-off between model performance and communication cost. Consequently, a significant body of research has focused on developing communication-efficient FL strategies. Early approaches centered on reducing communication frequency by allowing for more local computation on client devices before aggregation~\cite{mcmahan2017communication}. Another major line of work involves compressing the weight updates themselves through techniques like quantization~\cite{konevcny2016federated, alistarh2017qsgd}, which reduces the precision of weight updates, and sparsification~\cite{sattler2019robust}, which transmits only a subset of the most significant updates. 
More recently, sophisticated adaptive methods have been introduced. An example is FedLAMA~\cite{lee2023layer}, which employs a layer-wise adaptive aggregation strategy. Motivated by the observation that different layers converge at different rates, this approach adjusts the aggregation frequency of each layer based on its contribution to model discrepancy. However, these methods fundamentally assume a monolithic model architecture, and consequently, their approach to efficiency is restricted to the layer level. In contrast, our work is motivated by the functionally distinct, modular structures inherent in advanced robotics models. Instead of viewing the model as a simple stack of layers, we leverage the heterogeneous learning dynamics of its functional components. This allows us to propose a novel communication-efficient strategy tailored to multi-component systems, enabling a more granular and effective allocation of communication resources.


\begin{figure}[tbp]
    \centering
    \includegraphics[width=0.8\columnwidth]{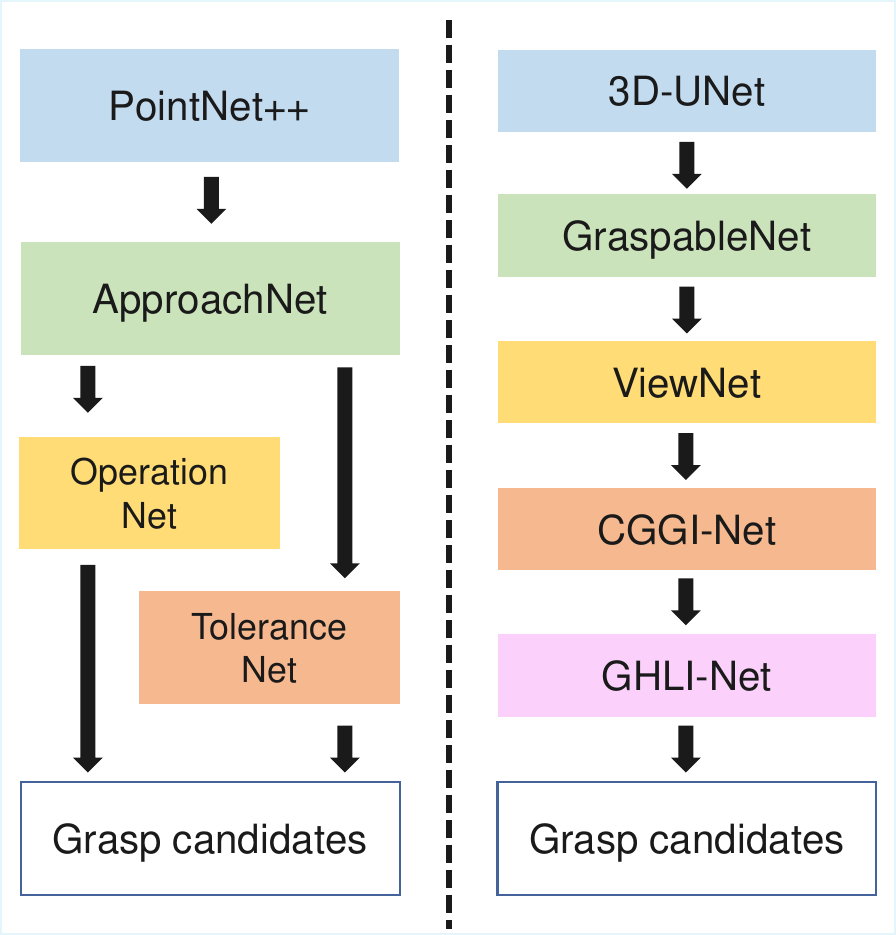} 
    \caption{Examples of multi-module architectures for GPD, from \cite{fang2020graspnet} (left) and \cite{wu2024economic} (right). These diagrams are simplified to show only learnable neural network modules, omitting other operations. The black arrows represent the main forward pass, though additional residual connections may exist in the actual models.}
    \label{fig:arch}
\end{figure}

\section{Preliminaries and Framework Overview}
This section introduces the background for our work. We start by explaining the standard FL framework and the 6-DoF grasp pose representation used for GPD. Next, we highlight a key opportunity for improvement that comes from the modular structure of GPD models. This opportunity is the motivation for the high-level overview of our proposed framework, which is presented at the end of the section.

\begin{figure}[tbp]
    \centering
    \includegraphics[width=\columnwidth]{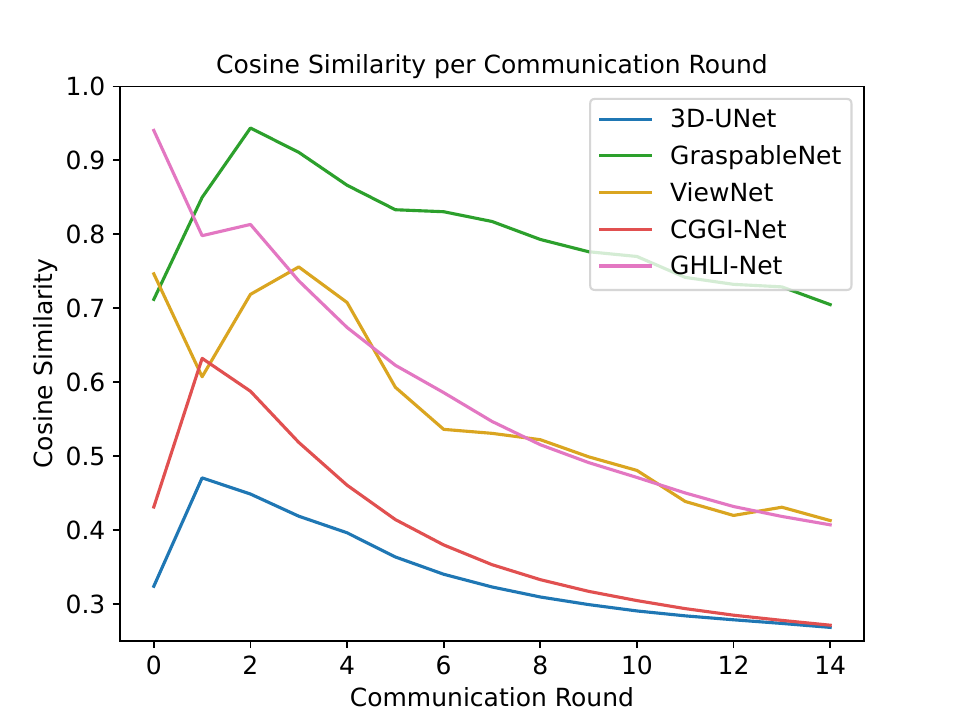}
    \caption{Cosine similarity between local client updates and the global update for different modules across communication rounds. The plotted values are averaged over three independent experiments, and the line colors are aligned with the corresponding modules in the right-hand architecture of Figure~\ref{fig:arch}. The plot reveals that distinct modules exhibit heterogeneous learning dynamics; for example, \texttt{GraspableNet} shows a higher similarity compared to \texttt{3D-UNet}.}
    \label{fig:cosine_analysis}
\end{figure}

\textbf{Federated Learning Framework.} FL is a decentralized machine learning paradigm for settings where training data is distributed across $n$ clients (indexed by $c \in \{1, \dots, n\}$) and cannot be centrally aggregated due to privacy or communication constraints. The objective is to collaboratively train a global model with parameters $w$ by minimizing a global loss function $L(w)$, which is the weighted average of each client's local loss $L_c(w; D_c)$ on their local dataset $D_c$:
$$\min_{w} L(w) = \sum_{c=1}^{n} k_c L_c(w; D_c)$$
The weight $k_c = |D_c|/|D|$ (where $|D| = \sum_{i=1}^n |D_i|$) typically represents the relative size of each client's dataset. The canonical algorithm to solve this objective is FedAvg~\cite{mcmahan2017communication}. The training process operates in communication rounds (indexed by $r$), where the server first distributes the current global model $w^r$ to all clients. Each client then performs local training to compute an updated model $w_c^{r+1}$ and, to preserve privacy, transmits only the model update $\Delta w_c^{r+1} = w_c^{r+1} - w^r$ back to the server. The server aggregates these updates to form the next global model $w^{r+1}$:
\[
w^{r+1} = w^r + \sum_{c=1}^{n} k_c \Delta w_c^{r+1}
\]

While more advanced optimizers exist~\cite{li2020federated, karimireddy2020scaffold}, we use FedAvg~\cite{mcmahan2017communication} as it provides a clear baseline for validating our module-wise strategy, which is orthogonal to the choice of optimizer.

\textbf{Grasp Pose Representation.} Directly learning the 6-DoF grasp pose matrix is challenging for neural networks due to the complex constraints of rotation matrices. Consequently, many state-of-the-art approaches~\cite{fang2020graspnet, wang2021graspness, wu2024economic} adopt a decoupled representation. Following the methodology in~\cite{wu2024economic}, we define a 6-DoF grasp $\mathbf{G}$ as a tuple $[\mathbf{c}, v, a, d, w, s]$. In this formulation, $\mathbf{c} \in \mathbb{R}^3$ is the grasp center point. The integers $v$, $a$, and $d$ represent the discretized approach direction, in-plane rotation, and grasp depth, respectively. Finally, $w \in \mathbb{R}$ is the grasp width and $s \in \mathbb{R}$ is the predicted grasp quality score.

\textbf{An Opportunity in Multi-Modular GPD Models for FL.} A key optimization opportunity in FL arises from the inherent structure of modern GPD models. As illustrated in Figure~\ref{fig:arch}, these models are typically not monolithic but are composed of multiple, functionally distinct neural network modules. Given that each module is specialized for a different sub-task, we posit that they will exhibit heterogeneous learning dynamics when trained under FL. For instance, some modules may converge faster, while others might require additional training.

\textbf{Overview of Our Proposed Framework.} Based on this premise, our work introduces a novel \textit{module-wise FL framework} designed to exploit these dynamics for communication-efficient learning. The core idea, which will be detailed in Section~\ref{sec:methodology}, is to supplement the standard full-model update with a second, resource-efficient phase focused only on a specific subset of modules. By strategically concentrating training and communication resources on specific components, our framework aims to maximize model performance for a given communication budget.


\section{Federated Learning Framework for \\ Grasp Pose Detection}
\label{sec:methodology}

\subsection{Analysis of Module-wise Learning Dynamics in FL}
\label{sec:similarity_analysis}
Modern deep learning models for GPD often feature multi-modular architectures~\cite{fang2020graspnet, wang2021graspness, wang2023granet, wu2024economic}, as illustrated in Figure~\ref{fig:arch}. \textcolor{black}{This modularity is a notable characteristic of robotics models, stemming from the inherent difficulty of training a neural network to interpret complex 3D spatial information \cite{fang2020graspnet}. Consequently, a common architectural approach is to employ a combination of multiple specialized modules.} For instance, the model presented in \cite{fang2020graspnet} consists of \texttt{PointNet++} \cite{qi2017pointnet++}, \texttt{ApproachNet}, \texttt{OperationNet}, and \texttt{ToleranceNet}. Similarly, the architecture in \cite{wu2024economic} is composed of \texttt{3D-UNet}, \texttt{GraspableNet}, \texttt{ViewNet}, \texttt{CGGI-Net}, and \texttt{GHLI-Net}. These modules are tasked with specific functions, such as extracting 3D features, determining approaching vectors, and estimating grasp quality. Motivated by this observation, we investigate whether these functionally distinct modules exhibit heterogeneous learning dynamics during FL training.

To quantify these dynamics, we analyze the alignment between local client updates and the aggregated global update for each module using cosine similarity. This metric has been employed in clustered FL research \cite{sattler2020clustered, ma2022convergence} to assess biases in client data. For each module $m \in M$, where $M$ is the set of all modules in the model, in communication round $r$, we compute the average module-wise similarity $S_m^r$ as:
\[S_m^r = \frac{1}{n} \sum_{c=1}^{n} \frac{\Delta w_{g,m}^r \cdot \Delta w_{c,m}^r}{\|\Delta w_{g,m}^r\| \|\Delta w_{c,m}^r\|}\]
Here, $\Delta w_{c,m}^r$ is the update for module $m$ from client $c$, and $\Delta w_{g,m}^r$ is the aggregated global update for that module ($\Delta w_{g,m}^r = \sum_{c=1}^n k_c \Delta w_{c,m}^r$).

Our analysis, conducted by applying FedAvg to a GPD model with a multi-modular architecture analogous to that in \cite{wu2024economic}, reveals heterogeneous learning dynamics at the module level. As illustrated in Figure~\ref{fig:cosine_analysis} (see Section~\ref{sec:experiments} for experimental details), these modules exhibit distinct similarity trajectories. For example, the module analogous to \texttt{GraspableNet} consistently shows high cosine similarity that diminishes gradually. Conversely, the module analogous to \texttt{3D-UNet} shows a lower initial similarity and a steady decline. These similarity values offer insight into the module's convergence state, showing the difference between learning \textit{general} and \textit{personalized} features, a key idea in~\cite{wang2022personalized, yi2024pfedmoe}. Specifically, if a module is far from its stationary point, it is primarily learning general features, leading to high similarity between local and global updates. As a module approaches convergence, however, it begins to learn local-data-specific features, causing the local updates to diverge and thus lowering their similarity to the global update. A low similarity suggests that the module is close to a stationary point, and thus can be considered fast-converging. In contrast, a module that maintains a high similarity value is interpreted as being relatively far from its stationary point, signifying a slow-converging module. This leads to our core design principle: to maximize the performance return on a given communication budget, additional training and aggregation efforts should be directed towards the slower-converging modules, as they are the primary bottlenecks to overall model performance.

\begin{algorithm}[!t]
    \caption{Module-wise FL (Server)}
    \label{alg:server_module_wise}
    \KwIn{Initial global model parameters $w^0$; 
    Total communication rounds $R$; 
    Number of clients $n$; 
    Client weights $\{k_c\}_{c=1}^n$;
    Local epochs $\tau_1, \tau_2$;
    \textcolor{black}{Number of slow-converging modules to select $\zeta$}}
    \KwOut{Final global model parameters $w^R$}
    Server distributes $w^{0}$ to all clients\;
    \For{\(r \leftarrow 0\) \KwTo \(R-1\)}{
        \ForPar{\(c \leftarrow 1\) \KwTo \(n\)}{ \label{alg:line:phase1_start}
            \(\Delta w_c \leftarrow \texttt{ClientUpdate}(c, M, \tau_1)\)\; \label{alg:line:phase1_end}
        } 
        $\Delta w_g \leftarrow \sum_{c=1}^{n} k_c \Delta w_c$\;
        \(w^{r+0.5} \leftarrow w^r + \Delta w_g\)\;
        \textcolor{black}{Calculate $S_m^r$ using updates from Phase 1\;} \label{alg:line:sim_calc}
        \textcolor{black}{\(\tilde{M}^r \leftarrow\) Select top-$\zeta$ modules with highest $S_m^r$\;} \label{alg:line:M-update}
        Server distributes $w^{r+0.5}$ \textcolor{black}{and $\tilde{M}^r$} to all clients\;
        \ForPar{\(c \leftarrow 1\) \KwTo \(n\)}{ \label{alg:line:phase2_start}
            \(\Delta w_c[\tilde{M}^{\textcolor{black}{r}}] \leftarrow \texttt{ClientUpdate}(c, \tilde{M}^{\textcolor{black}{r}}, \tau_2)\)\; \label{alg:line:phase2_end}
        }
        $\Delta w_g[\tilde{M}^{\textcolor{black}{r}}] \leftarrow \sum_{c=1}^{n} k_c \Delta w_c[\tilde{M}^{\textcolor{black}{r}}]$\;
        $w^{r+1} \leftarrow w^{r+0.5}$\;
        \(w^{r+1}[\tilde{M}^{\textcolor{black}{r}}] \leftarrow w^{r+0.5}[\tilde{M}^{\textcolor{black}{r}}] + \Delta w_g[\tilde{M}^{\textcolor{black}{r}}]\)\;
        Server distributes $w^{r+1}[\tilde{M}^{\textcolor{black}{r}}]$ to all clients\;
    }
    \Return{\(w^R\)}\;
\end{algorithm}

\subsection{Our Two-Phase Module-wise FL Algorithm}
Based on our core design principle that additional effort should be focused on slower-converging modules, we introduce a novel \textbf{two-phase module-wise FL algorithm}. In this framework, the set of slower-converging modules, $\tilde{M}$, is \textcolor{black}{adaptively identified within each communication round. This intra-round adaptation allows the algorithm to immediately focus on the modules that are the current learning bottlenecks.} Our framework, detailed in Algorithm~\ref{alg:server_module_wise}, is strategically designed to resolve the fundamental trade-off between full-model updates for robust generalization and partial-model updates for communication efficiency.

\textbf{Phase 1: Full-Model Training and Aggregation.}
As detailed in lines \ref{alg:line:phase1_start}-\ref{alg:line:phase1_end} of Algorithm~\ref{alg:server_module_wise}, \textcolor{black}{each round begins with clients training the full model ($M$) on their local data for $\tau_1$ epochs. The resulting updates, $\Delta w_c$, are transmitted to the server. The server then performs a weighted aggregation to compute the global update, $\Delta w_g$, and applies it to the current model ($w^r$) to produce an intermediate model, $w^{r+0.5}$. Crucially, the full-model updates collected in this phase provide a real-time snapshot of the learning dynamics across all modules.}

\textbf{Phase 2: \textcolor{black}{Adaptive} Module-Selective Training and Aggregation.}
Following Phase 1, the server immediately uses the collected updates ($\Delta w_c$) to calculate the module-wise cosine similarity $S_m^r$ for all modules. It then identifies the top-$\zeta$ modules with the highest similarity as the current set of slow-converging modules, $\tilde{M}^r$ (lines \ref{alg:line:sim_calc}-\ref{alg:line:M-update}). This newly identified set $\tilde{M}^r$, along with the intermediate model $w^{r+0.5}$, is then distributed to the clients.

In the second phase (lines \ref{alg:line:phase2_start}-\ref{alg:line:phase2_end}), clients apply the \texttt{ClientUpdate} procedure exclusively to this just-in-time identified subset $\tilde{M}^r$ for $\tau_2$ epochs. Clients transmit only the resulting partial updates, $\Delta w_c[\tilde{M}^r]$, back to the server. The server aggregates these to get the global partial update $\Delta w_g[\tilde{M}^{\textcolor{black}{r}}]$ and selectively applies it to form the final model for the round, $w^{r+1}$. For the next round, the server efficiently distributes only the updated partial model, $w^{r+1}[\tilde{M}^{\textcolor{black}{r}}]$, which clients use to reconstruct the full model locally. \textcolor{black}{This intra-round adaptive mechanism ensures that the additional training effort in Phase 2 is always directed at the most relevant modules based on the most current learning dynamics.}

In essence, our two-phase algorithm systematically separates the general, full-model update from the efficient, targeted refinement of slow-converging modules. This synergistic approach creates a more communication-efficient path to achieving high-performance GPD in a FL framework.


\section{Experiments}
\label{sec:experiments}

\begin{figure}[tbp]
\centering
\includegraphics[width=\columnwidth]{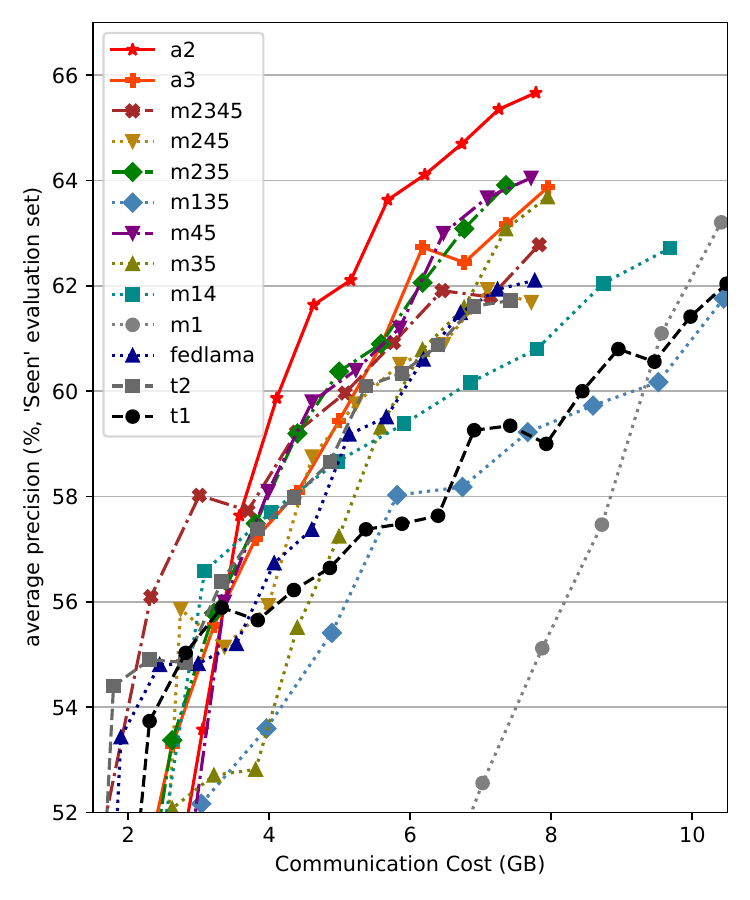}
\caption{Performance evaluation of methods on the \texttt{Seen} test set. This plot compares model accuracy against the total \textbf{communication cost} (in Gigabytes) for several methods.}
\label{fig:seen_eval}
\end{figure}

\subsection{Experimental Setup}
\textbf{Dataset and Model.}
Our experiments are conducted on the large-scale GraspNet-1B dataset \cite{fang2020graspnet}, using data captured by the Kinect sensor. Our implementation builds upon the off-the-shelf codebase of EconomicGrasp \cite{wu2024economic}. We chose this framework as it achieves state-of-the-art performance among frameworks that leverage \textit{less supervision}, making it an efficient baseline for our study. We adopt the standard architecture settings from the original work, with the feature dimensions of the 3D-UNet set to $[12, 24, 48, 96, 72, 72, 72, 72]$ and a fixed learning rate of $0.001$. All other hyperparameters and data processing steps follow the original implementation unless otherwise specified. All experiments were performed on a single NVIDIA RTX 4000 GPU.

\textbf{Federated Learning Protocol.}
To simulate a federated environment, we distributed the GraspNet-1B dataset across $n=20$ clients. In a single experimental run, a total of 100 unique scenes were used, with each client being randomly assigned data from 5 distinct scenes. To ensure the reliability of our results, we repeated all experiments 5 times.

\textbf{Evaluation Metrics and Cost Quantification.}
The performance metric is Average Precision (AP), calculated across a range of friction coefficients and averaged over the top-$k$ predictions (we set $k=10$). We utilize the standard \texttt{Seen}, \texttt{Unseen}, and \texttt{Novel} scene splits to assess generalization\textcolor{black}{, and report an \texttt{Overall} score calculated as the average AP across these three splits.} 

To quantify resource utilization, we define communication cost as the total volume of model parameters transmitted between the server and clients. For our proposed method, this can be expressed as: $$C_{\text{comm}} = 2nR(|w| + |w[\tilde{M}]|) + n|w|$$ where $|w|$ is the size of the full model parameters and $|w[\tilde{M}]|$ is the size of the module-wise updates. Computation cost is the cumulative GPU processing time for a single client \textcolor{black}{for the entire training process}.

\begin{table*}[t]
\centering
\begin{threeparttable}
\caption{Quantitative comparison of our proposed method with baselines and ablation studies. All evaluation results are averaged over at least five runs. The best evaluation results are marked in \textbf{bold}.}
\label{tab:main_results}
\begin{tabular}{c|c|cccc|c|c}
\hline
\multicolumn{2}{c|}{\multirow{2}{*}{\textbf{Method}}} &
\multicolumn{4}{c|}{\textbf{Evaluation Set (AP, \%)}} &
\multicolumn{1}{c|}{\multirow{2}{*}{\begin{tabular}{@{}c@{}}\textbf{\textcolor{black}{Communication}} \\ 
\textbf{\textcolor{black}{Cost (GB, $\downarrow$)}}\end{tabular}}} &
\multicolumn{1}{c}{\multirow{2}{*}{\begin{tabular}{@{}c@{}}\textbf{Computation} \\ 
\textbf{Cost (sec., $\downarrow$)}\end{tabular}}} \\ \cline{3-6}
\multicolumn{2}{c|}{} & \multicolumn{1}{c|}{\texttt{Seen} ($\uparrow$)} & \multicolumn{1}{c|}{\texttt{Unseen} ($\uparrow$)} & \multicolumn{1}{c|}{\texttt{Novel} ($\uparrow$)} & \multicolumn{1}{c|}{\texttt{Overall} ($\uparrow$)} & \multicolumn{1}{c|}{} & \multicolumn{1}{c}{} \\

\hline \hline
\multirow{4}{*}{Baseline}

& \texttt{\textcolor{black}{iso}} &
\multicolumn{1}{c|}{44.44 $\pm$ 4.50} & \multicolumn{1}{c|}{36.57 $\pm$ 2.58} & \multicolumn{1}{c|}{13.46 $\pm$ 1.77}
& \multicolumn{1}{c|}{31.49 $\pm$ 2.89} & \textcolor{black}{0} & 1000++ \\

& \texttt{t1} &
\multicolumn{1}{c|}{\textcolor{black}{62.09 $\pm$ 1.15}} & \multicolumn{1}{c|}{\textcolor{black}{45.29 $\pm$ 0.61}} & 
\multicolumn{1}{c|}{\textcolor{black}{17.92 $\pm$ 1.11}} & \textcolor{black}{41.76 $\pm$ 0.55} & 
\textcolor{black}{10.48} & \textcolor{black}{2365} \\

& \texttt{t2} &
\multicolumn{1}{c|}{\textcolor{black}{61.72 $\pm$ 0.91}} & \multicolumn{1}{c|}{\textcolor{black}{46.67 $\pm$ 1.13}} & 
\multicolumn{1}{c|}{\textcolor{black}{18.44 $\pm$ 0.65}} & \multicolumn{1}{c|}{\textcolor{black}{42.28 $\pm$ 0.67}} & 
\textcolor{black}{7.417} & \textcolor{black}{3295} \\

& FedLAMA~\cite{lee2023layer} &
\multicolumn{1}{c|}{\textcolor{black}{62.10 $\pm$ 0.56}} & \multicolumn{1}{c|}{\textcolor{black}{46.31 $\pm$ 1.45}} 
& \multicolumn{1}{c|}{\textcolor{black}{18.59 $\pm$ 1.10}} & \multicolumn{1}{c|}{\textcolor{black}{42.33 $\pm$ 0.90}} 
& \textcolor{black}{7.765} & \textcolor{black}{3121} \\

\hline \hline
\multirow{8}{*}{Ablation} & \texttt{m1} &
\multicolumn{1}{c|}{\textcolor{black}{63.20 $\pm$ 1.51}} & \multicolumn{1}{c|}{\textcolor{black}{44.21 $\pm$ 3.74}} 
& \multicolumn{1}{c|}{\textcolor{black}{17.65 $\pm$ 1.75}} & \multicolumn{1}{c|}{\textcolor{black}{41.69 $\pm$ 2.24}} 
& \textcolor{black}{10.41} & \multicolumn{1}{c}{\textcolor{black}{2718}} \\

& \texttt{m14} &
\multicolumn{1}{c|}{\textcolor{black}{62.71 $\pm$ 0.83}} & \multicolumn{1}{c|}{\textcolor{black}{46.32 $\pm$ 0.71}} 
& \multicolumn{1}{c|}{\textcolor{black}{19.08 $\pm$ 1.38}} & \multicolumn{1}{c|}{\textcolor{black}{42.70 $\pm$ 0.92}} 
& \textcolor{black}{9.683} & \multicolumn{1}{c}{\textcolor{black}{2324}} \\

& \texttt{\textcolor{black}{m35}} &
\multicolumn{1}{c|}{\textcolor{black}{63.68 $\pm$ 1.35}} & \multicolumn{1}{c|}{\textcolor{black}{48.37 $\pm$ 1.90}}
& \multicolumn{1}{c|}{\textcolor{black}{19.37 $\pm$ 1.70}} & \multicolumn{1}{c|}{\textcolor{black}{43.80 $\pm$ 1.64}}
& \textcolor{black}{7.947} & \multicolumn{1}{c}{\textcolor{black}{2688}} \\

& \texttt{\textcolor{black}{m45}} &
\multicolumn{1}{c|}{\textcolor{black}{64.04 $\pm$ 1.14}} & \multicolumn{1}{c|}{\textcolor{black}{49.40 $\pm$ 1.01}}
& \multicolumn{1}{c|}{\textcolor{black}{19.34 $\pm$ 0.417}} & \multicolumn{1}{c|}{\textcolor{black}{44.26 $\pm$ 0.83}}
& \textcolor{black}{7.714} & \textcolor{black}{2463} \\

& \texttt{\textcolor{black}{m135}} &
\multicolumn{1}{c|}{\textcolor{black}{61.75 $\pm$ 2.50}} & \multicolumn{1}{c|}{\textcolor{black}{44.47 $\pm$ 1.26}}
& \multicolumn{1}{c|}{\textcolor{black}{17.27 $\pm$ 0.70}} & \multicolumn{1}{c|}{\textcolor{black}{41.17 $\pm$ 1.46}}
& \textcolor{black}{10.44} & {\textcolor{black}{2455}} \\

& \texttt{m235} &
\multicolumn{1}{c|}{63.91 $\pm$ 1.17} & \multicolumn{1}{c|}{49.82 $\pm$ 1.35} 
& \multicolumn{1}{c|}{19.22 $\pm$ 1.62} & \multicolumn{1}{c|}{44.32 $\pm$ 1.38} 
& \textcolor{black}{7.358} & 2588 \\

& \texttt{\textcolor{black}{m245}} &
\multicolumn{1}{c|}{\textcolor{black}{61.69 $\pm$ 2.01}} & \multicolumn{1}{c|}{\textcolor{black}{48.19 $\pm$ 2.65}}
& \multicolumn{1}{c|}{\textcolor{black}{18.08 $\pm$ 1.58}} & \multicolumn{1}{c|}{\textcolor{black}{42.65 $\pm$ 1.84}}
& \textcolor{black}{7.717} & \textcolor{black}{2488} \\

& \texttt{m2345} &
\multicolumn{1}{c|}{\textcolor{black}{62.78 $\pm$ 0.41}} & \multicolumn{1}{c|}{\textcolor{black}{49.38 $\pm$ 1.56}} &
\multicolumn{1}{c|}{\textcolor{black}{19.31 $\pm$ 1.27}} & \multicolumn{1}{c|}{\textcolor{black}{43.82 $\pm$ 0.91}} & 
\textcolor{black}{7.828} & \textcolor{black}{2435} \\

\hline \hline
\multirow{2}{*}{Ours} & \texttt{\textcolor{black}{a3}} &
\multicolumn{1}{c|}{\textcolor{black}{63.17 $\pm$ 0.58}} & \multicolumn{1}{c|}{\textcolor{black}{50.11 $\pm$ 1.61}} &
\multicolumn{1}{c|}{\textcolor{black}{19.17 $\pm$ 1.18}} & \multicolumn{1}{c|}{\textcolor{black}{44.15 $\pm$ 0.69}} & 
\textcolor{black}{7.358} & \multicolumn{1}{c}{\textcolor{black}{2521}} \\

& \textcolor{black}{\texttt{a2}} &
\multicolumn{1}{c|}{\textcolor{black}{\textbf{64.42 $\pm$ 0.48}}} & \multicolumn{1}{c|}{\textcolor{black}{\textbf{51.89 $\pm$ 1.64}}} &
\multicolumn{1}{c|}{\textcolor{black}{\textbf{20.59 $\pm$ 1.48}}} & \multicolumn{1}{c|}{\textcolor{black}{\textbf{45.63 $\pm$ 0.87}}} &
\textcolor{black}{6.206} & \multicolumn{1}{c}{\textcolor{black}{2295}} \\

\hline
\end{tabular}

\begin{tablenotes}
\item[] \small \textcolor{black}{*The \texttt{iso} method represents isolated training and has zero communication cost by definition. Methods prefixed with \texttt{m} denote static ablations with pre-identified modules, while \texttt{a} denotes our adaptive approach where the number indicates the count of selected modules.}
\end{tablenotes}
\end{threeparttable}
    
\end{table*}

\subsection{Methods for Comparison}
Our investigation is based on the EconomicGrasp model, which we treat as a composition of five distinct modules: $m_1$ (\texttt{3D-UNet}), $m_2$ (\texttt{GraspableNet}), $m_3$ (\texttt{ViewNet}), $m_4$ (\texttt{CGGI-Net}), and $m_5$ (\texttt{GHLI-Net}). Our proposal is an adaptive two-phase algorithm that adaptively identifies and focuses on slow-converging modules. To validate its effectiveness, we compare it against several baselines and conduct extensive ablation studies. These studies include methods with pre-identified, static module sets for Phase 2. For all two-phase experiments, we set the local epochs to $\tau_1=1$ and $\tau_2=1$. 

\textbf{Baselines}:
\begin{itemize}
    \item \texttt{iso}: This baseline represents an \textit{isolated client} where each client trains a model using only its local data without any communication or aggregation. This establishes a performance lower bound, representing the outcome without collaboration.
    \item \texttt{t1} and \texttt{t2}: These represent standard FedAvg where the full model is trained in each communication round. We test with the number of local epochs set to $\tau = 1$ (\texttt{t1}) and $\tau = 2$ (\texttt{t2}) to evaluate the impact of additional local computation.
    \item FedLAMA \cite{lee2023layer}: An existing FL algorithm known for its layer-wise model aggregation strategy, included for a comprehensive comparison.
\end{itemize}

\textbf{Pre-identified $\tilde{M}$ (Ablation)}: These experiments test our hypothesis by using various pre-identified sets of modules ($\tilde{M}$) for Phase 2. This serves to validate the strategy of targeting specific modules and motivates our adaptive approach. Each variant is denoted by \texttt{m} followed by the indices of the chosen modules (e.g., \texttt{m14} uses the set $\tilde{M} = \{m_1, m_4\}$). The combinations tested in our ablation study are \texttt{m1, m14, m35, m45, m135, m235, m245}, and \texttt{m2345}.

\textbf{Proposed Method}: This is our main proposal, featuring the adaptive identification of slow-converging modules for Phase 2.  For clarity, we denote our adaptive methods as \texttt{a}$\zeta$, where $\zeta$ indicates the number of modules adaptively selected in each round. The set $\tilde{M}$ is adaptively chosen in each round based on the highest cosine similarity scores. Accordingly, \texttt{a2} and \texttt{a3} correspond to selecting the top $\zeta=2$ and $\zeta=3$ modules, respectively.

\begin{table*}[tbp]
\centering
\caption{\textcolor{black}{Comparison of grasp success rates for different FL methods in real-world experiments. The communication and computation costs for each method are identical to those reported in Table~\ref{tab:main_results}.}}
\label{tab:real_world_grasping}
\begin{tabular}{c *{6}{c}}
\toprule
\textbf{Method} & \textbf{\textcolor{black}{Scene 1}} & \textbf{\textcolor{black}{Scene 2}} & \textbf{\textcolor{black}{Scene 3}} & 
\textbf{\textcolor{black}{Scene 4}} & \textbf{\textcolor{black}{Scene 5}} & \textbf{\textcolor{black}{Overall}} \\
\midrule
\texttt{t1} & \textcolor{black}{70\% (10)} & \textcolor{black}{63\% (11)} & \textcolor{black}{70\% (10)} & 
\textcolor{black}{63\% (11)} & \textcolor{black}{62\% (13)} & \textbf{\textcolor{black}{65\% (55)}} \\

\texttt{\textcolor{black}{t2}} & \textcolor{black}{88\% (8)} & \textcolor{black}{63\% (11)} & \textcolor{black}{63\% (11)} & 
\textcolor{black}{70\% (10)} & \textcolor{black}{73\% (11)} & \textbf{\textcolor{black}{71\% (51)}} \\
 
\texttt{\textcolor{black}{m14}} & \textcolor{black}{88\% (8)} & \textcolor{black}{70\% (10)} & \textcolor{black}{58\% (12)} & 
\textcolor{black}{63\% (11)} & \textcolor{black}{67\% (12)} & \textbf{\textcolor{black}{68\% (53)}} \\

\texttt{m235} & \textcolor{black}{88\% (8)} & \textcolor{black}{70\% (10)} & \textcolor{black}{63\% (11)} & 
\textcolor{black}{63\% (11)} & \textcolor{black}{73\% (11)} & \textbf{\textcolor{black}{71\% (51)}} \\

\texttt{\textcolor{black}{m2345}} & \textcolor{black}{77\% (9)} & \textcolor{black}{77\% (9)} & \textcolor{black}{63\% (11)} & 
\textcolor{black}{70\% (10)} & \textcolor{black}{62\% (13)} & \textbf{\textcolor{black}{69\% (52)}} \\

\texttt{\textcolor{black}{a2}} & \textcolor{black}{88\% (8)} & \textcolor{black}{88\% (8)} & \textcolor{black}{63\% (11)} & 
\textcolor{black}{70\% (10)} & \textcolor{black}{67\% (12)} & \textbf{\textcolor{black}{73\% (49)}} \\
\bottomrule
\end{tabular}
\end{table*}

\begin{figure}[tbp]
    \centering
    \includegraphics[width=\columnwidth]{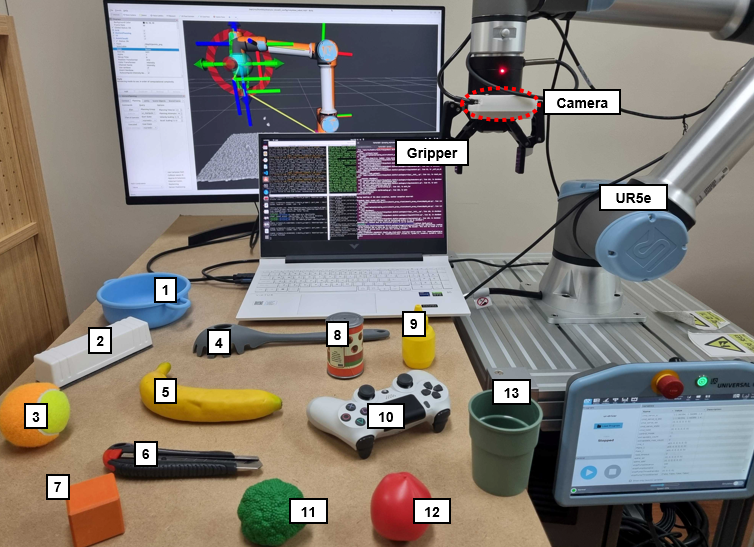}
    \caption{The experimental setup for real-world grasp validation. The robotic platform consists of a Universal Robots UR5e arm, a Robotiq 2F-85 parallel gripper, and a wrist-mounted Intel RealSense D435 camera. The \textcolor{black}{13} numbered objects,  were used to create cluttered test scenes.}
    \label{fig:robot_setup}
\end{figure}

\subsection{Results and Analysis}
Our extensive experiments demonstrate that the proposed adaptive two-phase algorithm, particularly \texttt{a2}, achieves a superior balance of performance and resource efficiency for federated GPD. As summarized in Table~\ref{tab:main_results}, our adaptive method \texttt{a2} achieves the highest \texttt{Overall} AP of 45.63\%, markedly outperforming all other approaches at a comparable communication budget. This trend is also clearly illustrated by the learning curve in Figure~\ref{fig:seen_eval}.

\textbf{The Necessity of Federated Learning.}
First, the poor performance of the \texttt{iso} (isolated client) baseline, with an \texttt{Overall} AP of only 31.49\%, confirms the necessity of FL. Without collaboration, models fail to generalize beyond their limited local data, underscoring the importance of a collaborative framework for developing robust GPD models.

\textbf{Superiority Over Standard FL Baselines.}
Our adaptive approach, \texttt{a2}, showed improved performance over the standard FL baselines. With an \texttt{Overall} AP of 45.63\%, it outperformed \texttt{t1} (41.76\%), \texttt{t2} (42.28\%), and FedLAMA (42.33\%). This performance gain was also achieved with greater resource efficiency. The \texttt{a2} method required 6.206 GB of communication and 2295s of computation, making it more efficient than all baselines: \texttt{t1} (10.48 GB, 2365s), \texttt{t2} (7.417 GB, 3295s), and FedLAMA (7.765 GB, 3121s). This result is notable as it suggests our method does not represent a typical accuracy-cost trade-off. Instead, it achieves a higher AP with a smaller resource budget, indicating that our adaptive strategy provides an efficient path for training high-performance models.

\textbf{The Critical Role of Adaptive Identification.}
A key insight from our experiments is the importance of adaptively identifying the slower-converging modules. While several pre-identified module sets in the ablation studies, such as \texttt{m45} and \texttt{m235}, performed well (44.26\% and 44.32\% \texttt{Overall} AP respectively), our adaptive \texttt{a2} method outperformed them, achieving the highest \texttt{Overall} AP of 45.63\%. This was also achieved with less communication overhead, as \texttt{a2} required only 6.206 GB, compared to 7.714 GB for \texttt{m45} and 7.358 GB for \texttt{m235}.

This performance gap points to a limitation of the static approach: the set of modules that act as a learning bottleneck is not fixed and can evolve during training. Static methods, which pre-identify a set $\tilde{M}$, rely on a fixed assumption about these learning dynamics. This can lead to inefficient resource allocation, as evidenced by the varied performance across the ablation studies. For instance, the \texttt{m1} and \texttt{m135} methods used more communication resources (10.41 GB and 10.44 GB) and similar levels of computation than the better-performing \texttt{m45} (7.714 GB), yet achieved lower accuracy (41.69\% and 41.17\% vs. 44.26\%). In contrast, our adaptive mechanism re-evaluates the learning dynamics in each communication round. This allows the additional training effort in Phase 2 to be directed toward the modules identified as most critical at that time, which explains the method's improved efficiency and performance.

In summary, these results show that our adaptive module-wise FL strategy offers an efficient path for training high-performance GPD models in a federated setting. By adaptively focusing updates on specific modules, our method achieves higher accuracy with lower computational and communication overhead compared to both standard FL methods and the static ablation approaches.

\subsection{Real-world Experiments}
To further validate the practical applicability and performance of our proposed FL framework, we conduct grasping experiments on a physical robotic system.

\textbf{Experimental Setup.} Our robotic setup consisted of a Universal Robots UR5e collaborative robot arm equipped with a Robotiq 2F-85 parallel gripper and an eye-in-hand Intel RealSense D435 depth camera for point cloud capture. A host PC running ROS2 served as the control station for communication and grasp execution. To create challenging grasping scenarios, we used a variety of common household and workshop objects, placing \textcolor{black}{7} to 8 of them in a cluttered arrangement for each experimental run. The complete setup and object set are depicted in Figure~\ref{fig:robot_setup}. The object sets are as follows: 
\textbf{Scene 1}: \{1, 2, 4, 9, 10, 11, 13\}; 
\textbf{Scene 2}: \{3, 5, 6, 7, 8, 11, 12\}; 
\textbf{Scene 3}: \{1, 2, 5, 8, 9, 11, 12\}; 
\textbf{Scene 4}: \{2, 3, 4, 5, 6, 7, 13\}; 
\textbf{Scene 5}: \{1, 2, 3, 4, 6, 8, 10, 13\}.

\textbf{Methods and Evaluation.} We compared the performance of \textcolor{black}{\texttt{t1}, \texttt{t2}, \texttt{m14}, \texttt{m235}, \texttt{2345}, and \texttt{a2}. The evaluation task is} to clear a cluttered scene of all its objects. For each run, we measured the total number of grasp attempts required to complete this task. A grasp was deemed successful if an object was securely lifted from the surface. \textcolor{black}{The success rate for each run was then calculated by dividing the number of objects in the scene by the total number of grasp attempts required to clear them.}

\textbf{Results and Discussion.} \textcolor{black}{The results of our real-world experiments, summarized in Table~\ref{tab:real_world_grasping}, are consistent with our simulation findings. The primary evaluation metric was the total number of grasp attempts required to clear each cluttered scene. The baseline \texttt{t1} model resulted in a 65\% success rate. Other methods, such as \texttt{t2} and \texttt{m235}, performed well, requiring 51 attempts to complete the task (71\% success rate). Our proposed \texttt{a2} method proved to be the most effective, clearing all scenes with the fewest attempts (49), which corresponds to the highest overall success rate of 73\%. This efficiency in grasp attempts suggests that the model trained with our adaptive strategy generates more reliable grasp poses, leading to fewer failures. These results support the practical applicability of our framework in a physical, cluttered environment.}

\section{Conclusion}
To address the high communication cost of FL on resource-constrained robots, we proposed \textcolor{black}{an adaptive} module-wise framework that reduces data transmission by selectively updating model components. \textcolor{black}{We validated our framework for GPD, where it outperformed standard and static FL methods in both accuracy and resource efficiency.} Since modular architectures are common in models for other robotics tasks, our approach has the potential to be applied more broadly as an efficient training strategy for other decentralized robotic systems.

\bibliography{reference}
\bibliographystyle{unsrt}
\end{document}